\documentclass[letterpaper, 10 pt, conference]{ieeeconf}  

\IEEEoverridecommandlockouts                              

\usepackage{amsmath}
\usepackage{graphicx} 
\usepackage{xcolor}
\usepackage{amssymb}
\usepackage[normalem]{ulem}
\usepackage{multirow}
\usepackage{booktabs}
\usepackage{balance}
\usepackage{makecell}
\usepackage{cite}
\usepackage[linesnumbered,ruled,vlined]{algorithm2e}
\SetAlgoSkip{smallskip}  

\title{\LARGE \bf
EmbodiedMind: Adaptive Data Curation and Prefix-Tree Reinforcement Learning for Efficient Embodied Intelligence
}

\author{
  Feifan Wang, Zongbing Zhang, Yu Zhang, Lingfeng Wang, Yurui Zhu, Jin Deng, \\
  Mingliang Zhang, Zhengguang Gao\textsuperscript{*}, Yongcheng Wang, Jin Xu, Ri Yang
  \thanks{All authors are with ZTE Corporation, Shenzhen, China. Corresponding author: Zhengguang Gao (gao.zhengguang@zte.com.cn)}
}

\begin{document}

\maketitle
\thispagestyle{empty}
\pagestyle{empty}

\begin{abstract}
Training embodied foundation models typically requires massive-scale datasets and extensive computational resources, yet often suffers from three critical limitations: (1) inefficient sample utilization due to low-informative samples; (2) imbalanced gradient contributions across heterogeneous tasks; and (3)  severe credit assignment problem in long-horizon planning, where trajectory-level rewards indiscriminately penalize all tokens. To address these issues, we propose an efficient training paradigm that achieves  state-of-the-art average performance through strategic data selection and hierarchical policy optimization. Our approach consists of three synergistic stages. First, Rejection Sampling-based Fine-Tuning \textbf{(RSFT)} filters out low-informative samples to establish robust behavioral priors while preventing distributional collapse. 
Second, Iterative Rejection GRPO \textbf{(IR-GRPO)} employs task-specific queues stratified by difficulty to keep datasets balanced across reinforcement learning iterations, coupled with a hybrid reward mechanism for precise cross-task feedback.
Third, to enhance long-horizon task planning, we introduce \textbf{Trie-GRPO}, a novel reinforcement learning algorithm based on action prefix trees, which enables step-level advantage estimation. This resolves the credit assignment problem by isolating intermediate correct decisions from downstream errors, while effectively balancing exploration efficiency and depth compared to conventional search trees.
As a result, \textbf{EmbodiedMind} achieves a state-of-the-art average performance of 70.02\% across 18 benchmarks, and significantly outperforms other embodied foundation models in long-horizon task planning accuracy. Our project will be released for reproducibility.
\end{abstract}
\section{Introduction}
\label{sec:introduction}
Embodied AI~\cite{sermanet2024robovqa,o2024open} represents the next frontier of artificial intelligence, aiming to integrate high-level cognitive capabilities with physical entities to enable agents to perceive, reason, and act autonomously in complex real-world environments. Fueled by the rapid advances in multimodal large language models~\cite{yang2025guiding,wu2025selp}, the field has shifted from narrow, single-task imitation learning toward general-purpose embodied foundation models. 

Despite significant progress in Embodied Intelligence, existing methods still face numerous challenges, which can be broadly categorized into three aspects.
First, conventional Supervised Fine-Tuning (SFT) and Reinforcement Learning (RL) protocols suffer from poor sample efficiency: they demand massive datasets and substantial computational resources, yet often converge to suboptimal policies because high-volume corpora contain abundant low-informative and high-variance trajectories. Second, existing dynamic sampling methods (e.g., DAPO~\cite{yu2026dapo}) filter zero-advantage samples without task-level calibration, leading to disproportionate discard of simple tasks and systematic neglect of difficult ones. This imbalance triggers policy degradation and induces a ``seesaw effect'' across capabilities. 
Third, standard Group Relative Policy Optimization (GRPO)~\cite{shao2024deepseekmath} applies trajectory-level advantages uniformly to all tokens, creating severe credit assignment issues in long-horizon planning where correct intermediate decisions are penalized for downstream errors. Tree-based search~\cite{tian2026seea,ding2026treegrpo,yang2025treerpo} offers a natural remedy by backpropagating values to individual nodes, but excessive exploration depth leads to combinatorial explosion.
To address these challenges, we propose an effective training paradigm for embodied foundation models, which couples three key components. First, to mitigate data inefficiency and task imbalance, we introduce a rejection sampling mechanism that partitions the data pool into task-specific queues and dynamically stratifies samples by estimated difficulty based on per-iteration model accuracy. By assembling balanced mini-batches across task types and difficulty tiers, this mechanism elevates the informational density of training data at the source. Second, to resolve the temporal credit assignment problem, we propose \textbf{Trie-GRPO}, a novel RL algorithm that leverages an \textit{action prefix tree} to evaluate step-level values rather than trajectory-level aggregates. This structure decouples intermediate decisions from downstream error propagation, thereby eliminating the uniform penalization inherent in standard GRPO. Compared with exhaustive tree search, Trie-GRPO achieves a principled balance between exploration depth and computational tractability, circumventing combinatorial explosion while preserving fine-grained credit discrimination. 
Finally, we integrate these components into a multi-stage iterative post-training pipeline with a multi-task reward mechanism, which jointly optimizes visual understanding, 2D/3D spatial perception, and long-horizon task planning.
Experiments show the proposed method improves training efficiency and performance over standard SFT and DAPO, enabling \textbf{EmbodiedMind} to achieve an average score of 70.02\% across 18 benchmarks and outperform strong same-scale embodied foundation model baselines.

In summary, our main contributions are as follows:
\begin{itemize}
    \item We propose an effective data curation and training paradigm that integrates task-aware dynamic difficulty stratification and a multi-stage iterative post-training pipeline with a hybrid reward mechanism.

    \item We introduce {Trie-GRPO}, a novel RL algorithm that uses an \textit{action prefix tree} to estimate step-level values instead of trajectory-level aggregates. 

    \item We comprehensively evaluate {EmbodiedMind} across open-source benchmarks, simulated environments, and real-world settings. Experimental results show that {EmbodiedMind} consistently outperforms strong baselines.
\end{itemize}

\section{Related Works}
\label{sec:related}

\subsection{Embodied Foundation Models}
Recent embodied foundation models have exhibited diverse capabilities. \textit{RynnBrain}~\cite{dang2026rynnbrain} employs spatio-temporal memory and physical reasoning via a mixture-of-experts architecture. \textit{RoboBrain2.5}~\cite{tan2026robobrain} enhances 3D spatial reasoning using depth-aware coordinates and dense temporal value estimation. \textit{MiMo-Embodied}~\cite{hao2025mimo} unifies embodied tasks and autonomous driving within a single vision-language model. \textit{EmbodiedBrain}~\cite{zou2025embodiedbrain} structures multimodal interactions into plans and actions with step-augmented optimization. \textit{Cosmos 3}~\cite{agarwal2026cosmos} introduces an omni-modal world model based on a mixture-of-Transformers architecture for unified physical AI. Despite advances in high-level cognition, 2D/3D perception, task supervision, and planning, these models remain heavily dependent on massive data and computation; RoboBrain 2.5, for instance, requires 12 million trajectories and 512 GPUs yet still struggles with the capability balance problem. In contrast, our approach mitigates this inefficiency via careful data curation strategy with task-specific queues and dynamic difficulty stratification, achieving balanced capability development and superior data efficiency.

\subsection{Training Algorithms for Embodied Foundation Models}
The training of embodied foundation models has recently accelerated through the adaptation of policy optimization techniques originally developed for large language models. GRPO stabilizes policy updates by computing advantages within response groups, while DAPO further applies dynamic sampling to filter zero-advantage samples, thereby prioritizing informative trajectories. Recent embodied research introduces domain-specific enhancements to these paradigms: \textit{Pelican-VL}~\cite{zhang2025pelican} improves data efficiency via Deliberate Practice Policy Optimization, which iteratively identifies weaknesses and performs targeted fine-tuning; \textit{EmbodiedBrain}~\cite{zou2025embodiedbrain} employs cold-start SFT with multimodal rejection sampling to establish foundational capabilities before RL optimization; and \textit{Robix}~\cite{fang2025robix} adopts modular training to decouple high-level reasoning from low-level execution. Despite these advances, existing protocols typically treat heterogeneous task types uniformly, leading to imbalanced gradient contributions and performance degradation in multi-task settings. Our proposed IR-GRPO resolves this issue by maintaining task-specific, difficulty-stratified FIFO (First In, First Out) queues, ensuring explicit data balance across heterogeneous task types during iterative rejection sampling.

\subsection{Long-Horizon Policy Optimization}
Multi-step tasks with delayed rewards require solving temporal credit assignment, motivating step-level verification to provide denser reward signals. \textit{Reinforced Reasoning}~\cite{wu2025reinforced} distills decision-making priors via SFT and optimizes a multimodal backbone with a rule-based, non-linear reward for action quality. \textit{REVER}~\cite{bo2025reinforced} trains RoboFarseer via RLVR, scoring plans with syntactic grammar checks and evaluating semantic coverage via ordered bipartite matching against ground-truth skill sequences. Alternatively, search-augmented policy optimization integrates search trees into the RL loop for backtracking on failures. \textit{SEEA-R1}~\cite{tian2026seea} combines MCTS~\cite{coulom2006efficient} with GRPO to backpropagate outcomes into step-wise Q-values, and incorporates a Multimodal Generative Reward Model for self-evolution without handcrafted rewards. However, rule-based methods rely on predefined gold plans that are expensive to annotate and assume one optimal solution per task, limiting generalization. Conversely, online tree-search approaches, while enabling step-wise backpropagation via MCTS, assume fully reversible environments and incur high search latency and computational overhead per expansion step. Our Trie-GRPO addresses these gaps by integrating group-relative advantage with action prefix trees for step-wise policy optimization in long-horizon planning, isolating early correct decisions from downstream errors without requiring simulation rollbacks or environment reversibility.

\begin{figure*}[] 
    \centering
    \includegraphics[width=1\linewidth]{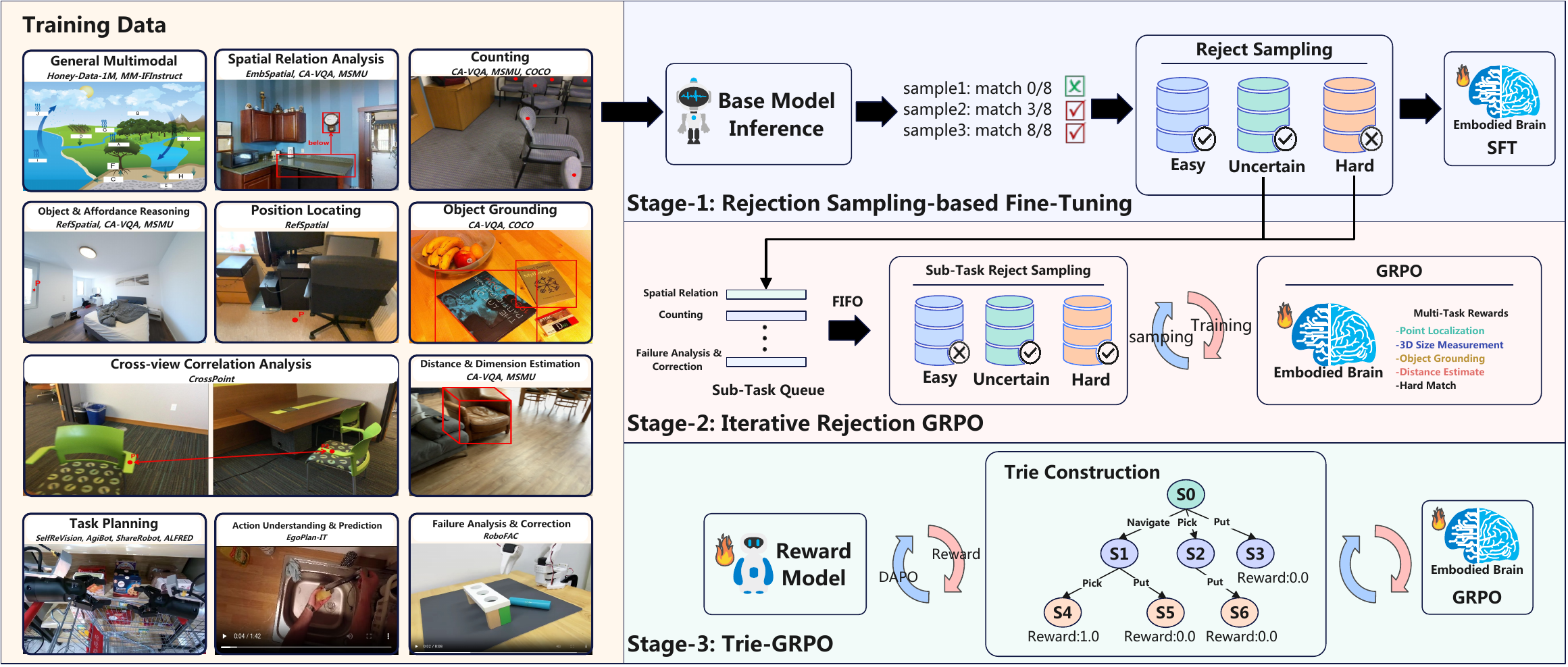} 
    \caption{The overall three-stage training framework of EmbodiedMind.} 
    \label{fig:framework} 
    \vspace{-10pt} 
\end{figure*}

\section{Method}
\label{sec:methodology}

In this section, we first describe the composition of the training data. Subsequently, we present our three-stage training methodology, comprising RSFT, IR-GRPO and Trie-GRPO. Through this three-stage training, we maximize embodied generalization capability with limited data scales while mitigating distributional bias during training. Our approach builds upon Qwen3-VL-8B~\cite{bai2025qwen3}, the same backbone adopted in RoboBrain 2.5~\cite{tan2026robobrain} and RynnBrain~\cite{dang2026rynnbrain}.

\subsection{Datasets}
\label{sec:datasets}

As illustrated in Fig.~\ref{fig:framework}, we construct a comprehensive multimodal training corpus with 11 capability-centric data types to ensure balanced skill acquisition.
General Multimodal~\cite{zhang2025bee,ding2025mm} data establish world knowledge and enable visually grounded instruction following. 
Spatial Relation and Cross-view Correlation data~\cite{du2024embspatial,daxberger2025mm,chen2026sd,wang2025towards} jointly develop egocentric, allocentric, and viewpoint-consistent reasoning. 
Counting and Object Grounding~\cite{lin2014microsoft,daxberger2025mm,chen2026sd} data enhance enumeration and localization accuracy in cluttered scenes, 
while Distance \& Dimension Estimation~\cite{lin2014microsoft,daxberger2025mm} data enable precise 3D coordinate regression and metric measurement. 
Object \& Affordance Reasoning and Position Locating tasks~\cite{lin2014microsoft,daxberger2025mm,zhou2026roborefer} bridge physical object properties with 2D coordinates through semantic affordance grounding. 
Task Planning data, including ALFRED~\cite{shridhar2020alfred}, SelfReVision~\cite{park2025making}, AgiBot~\cite{bu2025agibot_iros}, ShareRobot~\cite{ji2025robobrain}, are unified into a standard format composed of natural language responses, action-tagged plans, and structured action tuples. 
Action Understanding \& Prediction and Failure Analysis \& Correction data~\cite{chen2026egoplan,ye2025robofac} provide temporal reasoning and diagnostic capabilities. 
\subsection{Rejection Sampling-based Supervised Fine-Tuning}
\label{sec:rsft}
Full-dataset SFT is inefficient and converges to a suboptimal solution due to negligible gradient contributions from low-quality or already-learned samples. To address this, we propose RSFT, which uses rejection sampling to retain high-entropy samples, reducing computational costs while preventing early-stage distributional collapse.

The rejection sampling proceeds as follows: given the base model, we perform $k$ independent inferences at an elevated temperature to generate a candidate response set $\mathcal{Y}_{\text{cand}} = \{y_1, \ldots, y_k\}$ for each sample. 
These candidates are evaluated using the reward signals described in Section~\ref{sec:ir-grpo}, where continuous rewards are thresholded to determine correctness.
Based on matching outcomes, we stratify samples into three categories: \textit{Easy}, where all generations match the ground truth; \textit{Uncertain}, where only a subset of generations exhibit correctness; and \textit{Hard}, where no generation successfully matches. The final RSFT training set contains only uncertain samples and a controlled proportion of easy samples. Hard samples are excluded to prevent exposure to tasks far beyond the model's current capability during early training and minimize contamination from erroneous trajectories.

\subsection{Iterative Rejection GRPO}
\label{sec:ir-grpo}
Since post-RSFT models already exhibit preliminary spatial perception and planning, directly applying raw samples to RL introduces samples with negligible relative advantage, impairing training efficiency. We construct the RL data pool from uncertain and hard RSFT samples to provide high-information gradients. Although DAPO discards zero-advantage samples to filter trajectories, for difficult or long-tail tasks this induces a vicious cycle: the sampler expends substantial computation on samples that are ultimately rejected or only sparsely retained, while within a training batch the policy receives vanishingly few effective gradients for the tasks that most need improvement. To address this, we introduce two synergistic mechanisms: (1) EMA-based adaptive sampling quota and (2) hard-sample re-queueing.

We maintain separate FIFO queues for each task type and adopt iterative rejection sampling to prevent static data from rapidly losing its relative advantage once model capabilities evolve beyond the initial sampling distribution. At each iteration, we train with a subset of K samples and aim to balance the number of effective (i.e., non-zero-advantage) training samples across task types.

\textit{1) EMA-Based Adaptive Sampling Quota:} For task $j$ at iteration $t$, let $S_j^{(t)} \geq E_j^{(t)}$ be the sampled and effective sample counts. We define the instantaneous effective rate as $\eta_j^{(t)} = \frac{E_j^{(t)}}{S_j^{(t)}}$.  A low $\eta_j^{(t)}$ indicates that samples are difficult or underrepresented, signaling that the next iteration's quota should increase to maintain approximately $K/N$ effective samples per task, where $N$ is the number of task types. To avoid severe quota oscillations caused by single-batch variance, we apply an EMA (exponentially weighted moving average) smoother to suppress high-frequency noise and capture the underlying capability trend of each task: $\bar{\eta}_j^{(t)} = \beta \bar{\eta}_j^{(t-1)} + (1-\beta) \eta_j^{(t)}$ with $\beta = 0.8$. $\bar{\eta}_j^{(0)}$ is initialized by sampling the RSFT model on a small batch of data. This initialization bridges the two training stages and provides a reasonable prior on each task's initial difficulty for the RL stage. The adaptive quota for the next iteration is $\frac{K}{N \times \bar{\eta}_j^{(t)}}$.

\textit{2) Hard-Sample Re-queue:} Rather than discarding hard samples encountered in each iteration, we label and re-queue them into the corresponding task queue. When sampling, the sampler fills its quota by the following priority: it first draws fresh data; if insufficient, it falls back to previously sampled and re-queued hard samples, thereby improving data utilization.
As $\bar{\eta}$ decreases, the expanded sampling quota forces the policy to confront its weaknesses by replaying re-queued hard samples from difficult tasks, whereas easy tasks undergo no re-queueing.
Consequently, even a large-scale easy task contributes only a subset to training, whereas a small-scale difficult task is revisited multiple times until mastered or the re-queue limit $R_{\max}$ is reached, decoupling training intensity from the original data frequency.

We further design five reward mechanisms matched to output types to dynamically compute task-specific feedback signals: exact matching (0/1), semantic similarity (LLM judge, 0/1), point localization (normalized L1 distance), distance estimation (distance symmetry ratio), and 2D/3D box localization (F1-IoU joint reward), thereby providing dense, continuous rewards for numerical regression tasks.

\subsection{Trie-GRPO: Trie-Based Step-Level Reward Optimization}
\label{subsec:trie-grpo}

After two-stage training, EmbodiedMind achieves strong spatial reasoning and task understanding, yet its end-to-end planning remains suboptimal. We propose Trie-GRPO, which integrates GRPO with an Action Trie for step-level advantage estimation and fine-grained policy optimization, effectively addressing reward sparsity and credit assignment problems in complex multi-step tasks.

\subsubsection{Trie Grouping Mechanism}
\label{subsubsec:prefix-tree}

The core insight of Trie-GRPO is that the `group' for policy optimization should be defined conditionally at the \textbf{action prefix level}, rather than at the complete trajectory level. Specifically, for multiple trajectories sharing the same historical prefix $a_{1:t-1}$ (the action sequence from step 1 to $t-1$), their different action choices $a_t$ at step $t$ should form an independent decision sub-group, with relative advantage computation strictly confined within this sub-group.

We implement this hierarchical grouping mechanism by constructing an \textbf{Action Trie} structure, defined as follows:

\textbf{Node}: Each node in the tree corresponds to a unique action prefix sequence $v = (a_1, a_2, \ldots, a_t)$.

\textbf{Edge}: An edge from parent node $v_{\text{parent}} = a_{1:t-1}$ to child node $v_{\text{child}} = a_{1:t}$ is labeled with the action decision $a_t$.

\textbf{State Storage}: Each node $v$ maintains a state triplet $(Q, N, \mathcal{R})$, representing the estimated action value, visit count, and the set of cumulative discounted returns (i.e., $\mathcal{R}_v = \{R_{i,t} \mid \text{traj } i \text{ passes through } v \text{ after step } t\}$).

All $K$ trajectories sampled under the same task are inserted in parallel into a single Action Trie. For any trajectory's decision at step $t$, its relative advantage is determined not by the global reward of the entire trajectory, but exclusively by its \textbf{sibling child nodes} under the same parent prefix. This prefix tree-based local advantage computation mechanism offers the following core advantages over standard GRPO: (1) \textbf{Step-Level Credit Protection}: By comparing local Q-values of different trajectories under the same prefix, correct actions can still receive positive local advantage signals even when appearing in ultimately failed trajectories, avoiding ``collateral damage'' from subsequent erroneous decisions. (2) \textbf{Precise Branch Differentiation}: At decision divergence points, different action choices are directly compared as sibling nodes, ensuring correct actions receive positive advantages while incorrect actions receive negative advantages.


\subsubsection{Tree-Guided Signal Generation \& Automated Iterative Framework}
\label{subsubsec:training-signal}

Given an input query $q$, Trie-GRPO transforms raw rollout trajectories into advantage-annotated training data through six stages.

\textbf{Multi-candidate Trajectory Sampling}: Sample $K$ candidate trajectories $\{\tau_1, \ldots, \tau_K\}$ from the current policy $\pi_\theta$ for the input query $q$. Each trajectory $\tau_i$ is a sequence of natural language steps: $\tau_i = (s_{i,1}, s_{i,2}, \ldots, s_{i,T_i})$, where $s_{i,t}$ denotes the natural language description of step $t$, and $T_i$ is the total number of steps.

\textbf{Voting-based Reward Labeling}: For each trajectory $\tau_i$, we obtain $M$ independent evaluations from the reward model and assign a binary trajectory-level reward $r_i \in \{+1, -1\}$ via majority voting.

\textbf{Action Parsing and Trie Construction}: 
We parse each raw step $s_{i,t}$ into a canonical action tuple $a_{i,t} = (\text{action\_type}, \text{target\_obj}, \text{receptacle\_obj})$ to normalize semantic equivalents. An Action Trie $\mathcal{T}$ is then constructed by merging shared action prefixes across all parsed trajectories.

\textbf{Temporal-Discounted Return Propagation}: We propagate the trajectory-level reward $r_i$ back along the trie path with a discount factor $\gamma$, computing the $Q$-value of node $v$ as the mean of discounted returns:
\begin{equation}
R_{i,t} = \gamma^{T_i - t} \cdot r_i, \quad Q(v) = \frac{1}{N_v} \sum_{i \in \mathcal{I}(v)} R_{i,t}
\end{equation}
where $\mathcal{I}(v)$ denotes the set of trajectory indices passing through node $v$.

 \textbf{Sibling-Normalized Advantage Computation}: For each parent node $u$ with children set $\mathcal{C}_u = \{v_1, \ldots, v_k\}$, we compute the normalized advantage for each child action:
\begin{equation}
\label{eq:group-advantage}
\begin{gathered}
\mu_u = \frac{1}{k} \sum_{j=1}^{k} Q(v_j), \quad
\sigma_u = \sqrt{\frac{1}{k} \sum_{j=1}^{k} (Q(v_j) - \mu_u)^2 + \epsilon}, \\[1ex]
A(v_j) = \frac{Q(v_j) - \mu_u}{\sigma_u}
\end{gathered}
\end{equation}

\textbf{Low-Variance Sample Filtering}: We filter out samples with near-zero advantages ($|A(v)| \leq \delta$, e.g., $\delta=0.001$) to remove uninformative gradients. The final training dataset is constructed as a set of triplets $\mathcal{D} = \{((q, s_{1:t-1}), s_t, A(a_{1:t}))\}$.

Based on the computed step-level advantages, we define the \textbf{action-level} importance sampling ratio for policy update. The loss is computed exclusively over the tokens corresponding to $a_t$, with all other context tokens masked out:
\begin{equation}
\rho_{s_t}(\theta) = \frac{\pi_\theta({o_i \in s_t} \mid q,s_{1:t-1})}{\pi_{\theta_{\text{old}}}({o_i \in s_t} \mid q,s_{1:t-1})}
\end{equation}
The Trie-GRPO loss is then formulated as:


\begin{equation}
\begin{split}
\mathcal{L}(\theta) = &- \frac{1}{N_u} \sum_{v \in \mathcal{C}_u} \Big\{ \min \big[ \rho_{s_t}(\theta) A(v), \\
& \quad \mathrm{clip}(\rho_{s_t}(\theta), 1\pm \epsilon) A(v) \big] \\
& \quad - \beta D_{\mathrm{KL}}[\pi_\theta(s_t) \parallel \pi_{\mathrm{ref}}(s_t)] \Big\}
\end{split}
\end{equation}
where $\mathcal{C}_u$ denotes the set of all children of node $u = a_{1:t-1}$, 
$\epsilon$ is the clipping hyperparameter, $\beta$ controls the KL penalty strength, and $D_{\text{KL}}$ denotes the KL divergence on raw step $s_t$ between the reference model $\pi_{\text{ref}}$ and policy model $\pi_{\theta}$.

\begin{figure}[t]
    \centering
    \includegraphics[width=\linewidth]{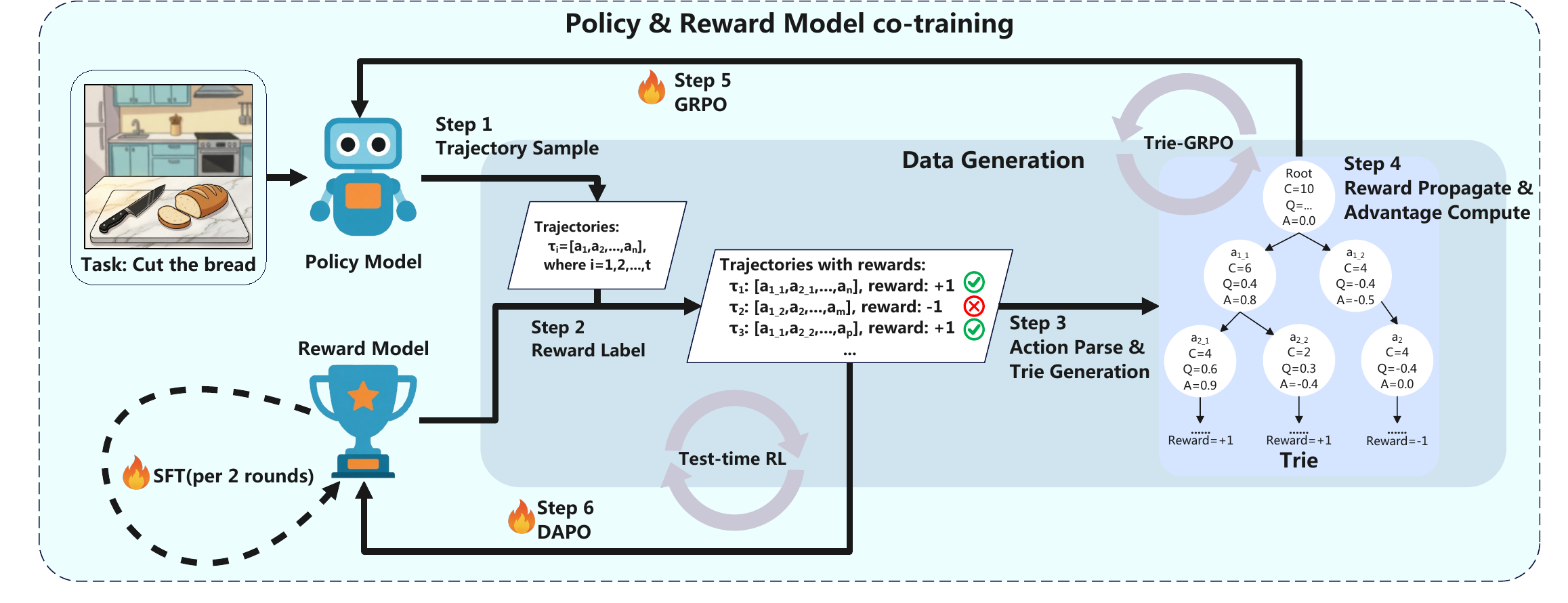}
    \vspace{-5pt}
    \caption{Overview of the Trie-GRPO automated iterative training framework. The framework partitions tasks into batches and iteratively executes data generation and model co-evolution.}
    \label{fig:trie-grpo-framework}
    \vspace{-10pt}
\end{figure}

As shown in Fig.~\ref{fig:trie-grpo-framework}, we embed Trie-GRPO into a multi-round automated training framework that enables progressive co-optimization of the policy and reward model through iterative data generation and model updates.

\section{Experiments}
\label{sec:experiments}


\subsection{Training Hyperparameters}
For RSFT, the training data volume is 349K with a batch size of 16.
For IR-GRPO, the training data volume is 800 per round with a batch size of 64 and $G=8$.
For Trie-GRPO, we set $K=15$ for the policy model, generating 15 candidate trajectories per task, with 40 tasks trained per round.
The group size of the reward model is set to $G=5$, generating 5 candidate rewards per trajectory, with 500 trajectories trained per round.
All model training and data processing are conducted on 8$\times$H800 GPUs, with one additional GPU required for rollout in reinforcement learning.

It is worth noting that the group size $k$ in Eq.~(\ref{eq:group-advantage}) is not fixed but dynamically determined by the policy model under the current prefix based on action uncertainty.
For instance, the average group size for the `slice' prefix is 3.52 (involving subsequent actions such as washing or placing the knife), whereas the average group size for the `open' prefix is only 2.03 (involving subsequent actions such as placing or picking up objects). On average, each trie contains 6.06 branching nodes and 10.26 leaf nodes, resulting in an average group size of 2.53.

\subsection{Evaluation Benchmarks}
\label{subsec:benchmark_descriptions}
We comprehensively evaluate EmbodiedMind on benchmarks in four categories: general multimodal understanding (AI2D~\cite{kembhavi2016diagram}, MMSTAR~\cite{chen2024we}, MMMU~\cite{yue2024mmmu}, and OCRBench~\cite{liu2024ocrbench}), 2D spatial reasoning (CV-Bench~\cite{tong2024cambrian}, RoboSpatial~\cite{song2025robospatial}, RefSpatial~\cite{zhou2026roborefer}, EMbSpatial~\cite{du2024embspatial}, ERQA~\cite{team2025gemini}, and CrossPoint~\cite{wang2025towards}), 3D spatial measurement (MSMU~\cite{chen2026sd}, BLINK~\cite{fu2024blink}, and Q-Spatial~\cite{liao2024reasoning}), and task planning (ShareRobot-based step validity judgment and next-step prediction~\cite{ji2025robobrain}, EgoPlan-Bench~\cite{chen2026egoplan,qiu2026egoplan} and RoboBench~\cite{luo2025robobench}) Additionally, we assess long-horizon planning on VLM-PlanSim-99~\cite{zou2025embodiedbrain}, an AI2-THOR-based simulated benchmark comprising 11 manipulation actions and one navigation action, with tasks requiring up to 25 planning steps and an average trajectory length of 10.73.

\begin{table*}
\centering
\caption{Evaluation results across different benchmarks. Red indicates the best result; underlined indicates the second-best result. (EB-7B: EmbodiedBrain-7B~\cite{zou2025embodiedbrain}, Qwen3-8B: Qwen3-VL-8B~\cite{bai2025qwen3}, RB-8B: RoboBrain2.5-8B-NV~\cite{tan2026robobrain}, MiMo-7B: MiMo-Embodied-7B~\cite{hao2025mimo}, Rynn-8B: RynnBrain-8B~\cite{dang2026rynnbrain}, Pelican-7B: Pelican1.0-VL-7B~\cite{zhang2025pelican}, Cosmos3: Cosmos3-Nano~\cite{agarwal2026cosmos})}
\label{tab:main_results}
\begin{tabular}{ccccccccc} 
\hline
\textbf{Benchmark} & \textbf{EB-7B}         & \textbf{Qwen3-8B}               & \textbf{RB-8B}                  & \textbf{MiMo-7B}       & \textbf{Rynn-8B}       & \textbf{Cosmos3}    & \textbf{Pelican-7B}     & \textbf{Ours-8B}                 \\ 
\hline
\multicolumn{9}{c}{\textbf{General Benchmarks}}                                                                                                                                                                               \\ 
\hline
AI2D~              & 82.61                  & \textcolor{red}{84.84}          & 82.45                           & 82.90                  & \uline{84.26}          & 83.87        & 83.42          & 82.51                            \\
MMSTAR~            & 62.17                  & \textcolor{red}{69.93}          & 67.00                           & \uline{68.27}          & 65.47                  & 64.40            & 62.33      & 67.27                            \\
MMMU~              & 52.67                  & \uline{66.48}                   & 56.90                           & \textcolor{red}{68.76} & 59.14                  & 55.71           &55.33       & 63.34                            \\
OCRBench~          & 78.30                  & \textcolor{red}{81.40}          & 78.10                           & 67.60                  & 72.80                  & 75.60            &76.50      & \uline{78.50}                    \\
\textbf{Average}   & \textbf{68.94}         & \textcolor{red}{\textbf{75.66}} & \textbf{71.11}                  & \textbf{71.88}         & \textbf{70.42}         & \textbf{69.90}     & \textbf{69.39}     & \textbf{\uline{72.91}}           \\ 
\hline
\multicolumn{9}{c}{\textbf{2D Reasoning}}                                                                                                                                                                                     \\ 
\hline
CV-Bench~          & 80.67                  & 87.06                           & 87.87                           & 56.62                  & \uline{87.90}          & \textcolor{red}{88.41}   & 81.01  & 87.05                            \\
CrossPoint~        & 17.30                  & 29.20                           & \textcolor{red}{75.60}          & 31.70                  & 38.60                  & 37.40     & 24.60             & \uline{72.70}                    \\
RoboSpatial~       & 41.14                  & 61.43                           & 68.57                           & 58.86                  & \textcolor{red}{72.29} & 58.29         & 55.71         & \uline{69.43}                    \\
RefSpatial~        & 0.25                   & 30.75                           & \textcolor{red}{62.62}          & 36.22                  & 58.00                  & \uline{61.00}     & 37.00      & 56.00                            \\
EMbSpatial~        & 74.70                  & 78.72                           & 75.06                           & 77.70                  & \uline{80.88}          & \textcolor{red}{81.26}    & 72.31  & 78.08                            \\
ERQA~              & 41.25                  & 42.25                           & \uline{45.00}                   & 40.75                  & 41.00                  & \textcolor{red}{48.25}  & 40.25  & 44.00                            \\
\textbf{Average}   & \textbf{42.55}         & \textbf{54.90}                  & \textcolor{red}{\textbf{69.12}} & \textbf{50.31}         & \textbf{63.11}         & \textbf{62.44}   & \textbf{51.81}      & \textbf{\uline{67.88}}           \\ 
\hline
\multicolumn{9}{c}{\textbf{3D Measurement}}                                                                                                                                                                                   \\ 
\hline
MSMU~              & 33.53                  & 33.79                           & \uline{66.96}                   & 29.38                  & 51.91                  & 49.70        & 55.21           & \textcolor{red}{85.81}           \\
BLINK~             & \textcolor{red}{87.41} & 79.72                           & 78.32                           & 69.93                  & 74.13                  & 79.72           & 80.42        & \uline{84.62}                    \\
Q-Spatial~         & 38.38                  & 59.78                           & \textcolor{red}{70.85}          & 60.89                  & 49.82                  & 61.99       & 39.85            & \uline{68.63}                    \\
\textbf{Average}   & \textbf{53.11}         & \textbf{57.76}                  & \textbf{\uline{72.04}}          & \textbf{53.40}         & \textbf{58.62}         & \textbf{63.80}    & \textbf{58.49}     & \textcolor{red}{\textbf{79.69}}  \\ 
\hline
\multicolumn{9}{c}{\textbf{Planning}}                                                                                                                                                                                         \\ 
\hline
ShareRobot-Judge~  & 90.65                  & 95.55                           & 94.60                           & 91.30                  & 67.00                  & \uline{96.00}   & 92.80      & \textcolor{red}{96.35}           \\
ShareRobot-Choice~ & 81.18                  & 75.92                           & 81.44                           & \textcolor{red}{90.68} & 82.55                  & 76.73         & 80.75         & \uline{83.92}                    \\
EgoPlan-1~         & 49.13                  & 48.68                           & 45.28                           & 45.46                  & \textcolor{red}{53.83} & 49.16      & 44.65           & \uline{51.29}                    \\
EgoPlan-2          & \uline{50.42}          & 47.00                           & 48.06                           & 42.36                  & \textcolor{red}{54.98} & 45.55         & 41.06         & 47.38                            \\
RoboBench~         & 37.05                  & 39.25                           & 39.00                           & \textcolor{red}{45.06} & 34.51                  & 40.25          & 31.28        & \uline{43.42}                    \\
\textbf{Average}   & \textbf{61.69}         & \textbf{61.28}                  & \textbf{61.68}                  & \textbf{\uline{62.97}} & \textbf{58.57}         & \textbf{61.54}   & \textbf{58.11}     & \textcolor{red}{\textbf{64.48}}  \\ 
\hline
\textbf{Overall}   & \textbf{55.49}         & \textbf{61.76}                  & \textbf{\uline{67.98}}          & \textbf{59.14}         & \textbf{62.73}         & \textbf{64.07}   & \textbf{58.58}       & \textbf{\textcolor{red}{70.02}}  \\ 
\hline
\multicolumn{9}{c}{\textbf{End-to-end Sim}}                                                                                                                                                                                   \\ 
\hline
VLM-PlanSim-99~    & 31.31         & 13.13                           & 25.25                           & 19.19                  & 11.11           & \uline{36.36}       & 1.01                   & \textcolor{red}{66.67}           \\
\hline
\end{tabular}
\end{table*}

\subsection{Evaluation Results}
\label{subsec:evaluation_results}



We conducted comprehensive evaluations against state-of-the-art multimodal and embodied foundation models at comparable parameter scales. As shown in Tab.~\ref{tab:main_results}, \textbf{EmbodiedMind} achieves an superior average performance of \(70.02\%\) across 18 benchmarks, surpassing prominent baselines including Qwen3-VL-8B and RoboBrain2.5-8B, \textit{although using merely 522K training samples compared to 12.4M for RoboBrain2.5-8B}. On the VLM-PlanSim-99 benchmark, our method further improves long-horizon planning by 30.31\% over the existing state-of-the-art.

\subsection{Ablation Study}

\textbf{Efficiency of Training.} Tab.~\ref{tab:time_analy} reports the training efficiency of our three-stage method. For Stage 1, SFT denotes direct training on 50k samples. RSFT, in contrast, first performs rejection sampling and then trains on the retained Uncertain and Easy samples. To isolate the effect of rejection sampling, RSFT is evaluated separately on two datasets: 50k Cross-point and 50k Honey-1M. Cross-point comprises structured point localization data, for which a rule-based matching strategy yields high inference efficiency. Honey-1M, conversely, consists of unstructured general multimodal data, whose rejection sampling relies on LLM scoring and is thus less efficient. RSFT takes 65 minutes for 50k Cross-point (56\% of which is rejection sampling), versus 110 minutes for Honey-1M (85\% of which is rejection sampling), indicating that Stage 1 training efficiency is dominated by rejection sampling. For Stage 2, the efficiency difference mainly arises from RL training: one IR-GRPO round takes only 155 minutes (14.5\% for rejection sampling), whereas DAPO takes 352 minutes on the same data volume, more than twice that of IR-GRPO, demonstrating improved training efficiency. For Stage 3, each Trie-GRPO round takes 58.76 minutes, with an additional 22.29\% relative overhead mainly from trie construction and advantage computation.

\begin{table*}[h]
\centering
\caption{Comparison of Training Time Across Three Stages}
\label{tab:time_analy}
\begin{tabular}{cccccccc}
\toprule
\textbf{Training Stage} & \multicolumn{3}{c}{\textbf{Stage 1 (min/50k)}} & \multicolumn{2}{c}{\textbf{Stage 2 (min/round)}} & \multicolumn{2}{c}{\textbf{Stage 3 (min/round)}} \\
\cmidrule(lr){2-4} \cmidrule(lr){5-6} \cmidrule(lr){7-8}
Training Method & SFT & RSFT (Cross-point) & RSFT (Honey-1M) & DAPO & IR-GRPO & GRPO & Trie-GRPO \\
\midrule
Training Time & 79 & 65 & 110 & 352 & 155 & 48 & 58 \\
\bottomrule
\end{tabular}
\vspace{-10pt}
\end{table*}

\begin{figure}[t]
    \centering
    \includegraphics[width=\linewidth]{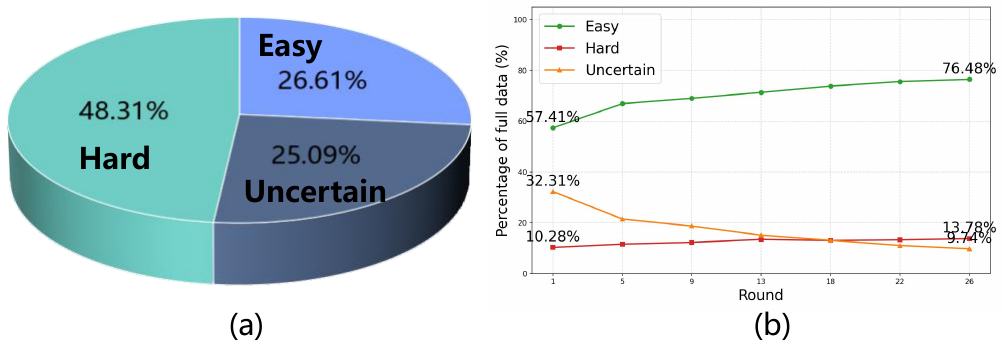}
    \caption{Training samples distribution after rejection sampling. (a) Pie chart of the sample distribution in the RSFT stage; (b) Line chart of the sample distribution across training rounds in IR-GRPO.}
    \label{fig:pancake_fig}
    \vspace{-20pt}
\end{figure}

We also present two separate visualizations: the training sample distribution after rejection sampling in RSFT and its variation across training rounds in IR-GRPO. As shown in Fig.~\ref{fig:pancake_fig}, hard samples decrease substantially after RSFT. For IR-GRPO, as iterations increase, Uncertain samples drop from 32.31\% to 9.74\%, while easy samples rise from 57.41\% to 76.48\%. This indicates that our method transforms samples with weak answer consistency into accurately solvable easy samples, significantly improving model accuracy.

\textbf{Effectiveness of the Training Strategy.} We first evaluate the effectiveness of RSFT against standard SFT with random sampling. As shown in Tab.~\ref{tab:sft_ablation_results}, RSFT achieves superior performance across all four capabilities. 
These results indicate that standard SFT is adversely affected by noisy, low-quality samples that lead to suboptimal policy performance, whereas RSFT's strategic data curation ensures both higher data quality and smaller distribution shift.

\begin{figure}[t]
    \centering
    \includegraphics[width=\linewidth]{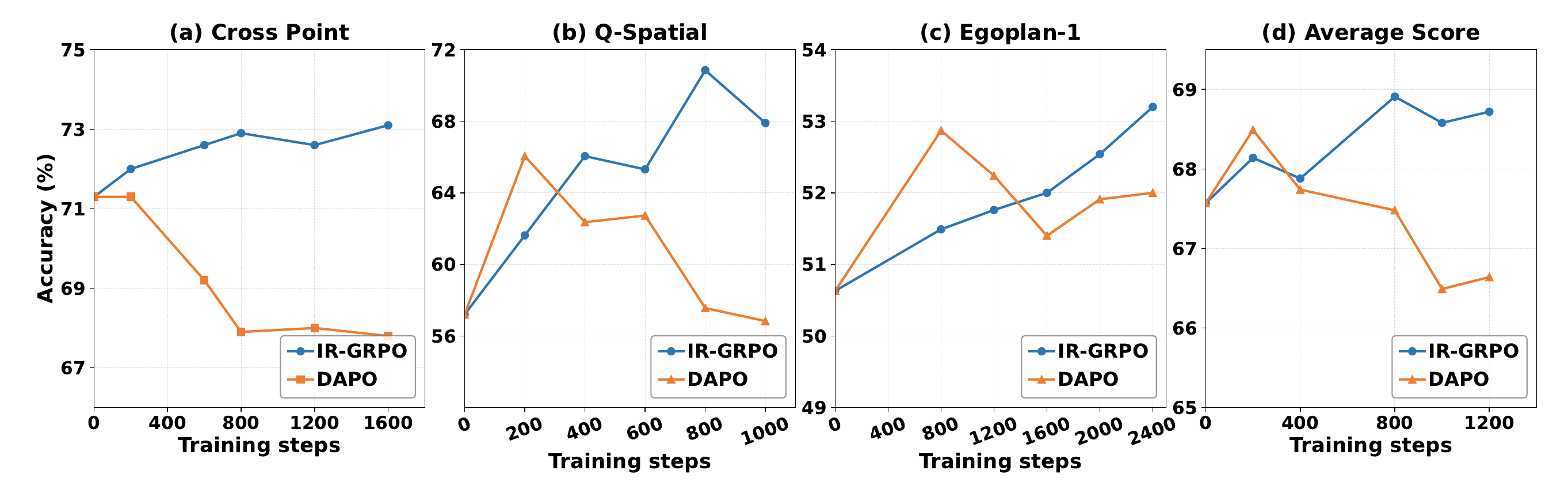}
    \caption{Comparison of DAPO and IR-GRPO on Key Benchmarks. The Average Score is Computed Across 14 Benchmarks Spanning 2D Reasoning, 3D Measurement, and Planning Categories.}
    \label{fig:ablation_dapo}
\end{figure}

For Stage 2, we compare IR-GRPO against DAPO. As detailed in Tab.~\ref{tab:sft_ablation_results}, IR-GRPO demonstrates superior stability and generalization compared to DAPO. 
We further evaluate both methods on three representative benchmarks: CrossPoint, Q-Spatial, and EgoPlan-1, as well as on the average score over 14 benchmarks covering 2D reasoning, 3D measurement, and planning. 
As illustrated in Fig.~\ref{fig:ablation_dapo}, IR-GRPO consistently outperforms DAPO across all evaluated dimensions.
DAPO exhibits an initial performance increase followed by severe degradation on spatial reasoning tasks after 200 steps; for instance, its Q-Spatial accuracy drops precipitously from 66.05\% at step 200 to 56.83\% at step 1000.
In contrast, IR-GRPO maintains stable and consistent improvement across these benchmarks, indicating that our approach preserves task-level balance throughout training.
\begin{table}[]
\centering
\caption{Training Results Across Stages 1 and 2}
\label{tab:sft_ablation_results}
\begin{tabular}{ccccc}
\toprule
\multirow{2}{*}{\textbf{Evaluation Dimension}} & \multicolumn{2}{c}{\textbf{Stage-1}} & \multicolumn{2}{c}{\textbf{Stage-2}} \\
\cmidrule(lr){2-3} \cmidrule(lr){4-5}
                                  & \textbf{SFT}     & \textbf{RSFT}     & \textbf{DAPO}   & \textbf{IR-GRPO}   \\
\midrule
General Benchmarks                & 68.44            & 72.76             & 73.51           & 73.34              \\
2D Reasoning                      & 65.06            & 67.49             & 66.07           & 67.67              \\
3D Measurement                    & 72.66            & 74.39             & 69.50           & 78.93              \\
Planning                          & 63.17            & 63.57             & 65.18           & 64.37              \\ \hline
\textbf{Overall}                  & \textbf{66.55}    & \textbf{68.72}   & \textbf{68.04}   & \textbf{69.89} \\
\bottomrule
\end{tabular}
\vspace{-10pt}
\end{table}

To demonstrate the effectiveness of Trie-GRPO, we compare it against GRPO on VLM-PlanSim-99. To account for inference noise, we report the mean and variance over five runs. As shown in Tab.~\ref{tab:VLM-PlanSim-99}, our method outperforms GRPO by 4.83\% and exceeds the Stage 2 baseline by 8.16\%.

\begin{table}[ht]
  \centering
  \caption{comparison of stage 3 algorithm across five inference seeds}
  \begin{tabular}{cccc}
    \hline
    \multirow{2}{*}{\textbf{Evaluation Metric}} & \multirow{2}{*}{\textbf{Stage2}} & \multicolumn{2}{c}{\textbf{Stage-3}} \\
    \cline{3-4}
      & & \textbf{GRPO} & \textbf{Trie-GRPO} \\
    \hline
    \textbf{Avg$\pm$Std} & 58.17 $\pm$ 0.41 & 61.50 $\pm$ 0.84 & 66.33 $\pm$ 0.82 \\
    \hline
  \end{tabular}
  \label{tab:VLM-PlanSim-99}
  \vspace{-5pt}
\end{table}

We also compare our method with SEEA-R1~\cite{tian2026seea} on the ALFWorld-thor test-unseen split. 
Despite sharing the idea of providing fine-grained rewards for individual action steps,
they differ fundamentally: SEEA-R1 adopts step-wise ReAct reasoning with per-step error recovery, whereas Trie-GRPO generates long-horizon plans and replans upon failure. For the experimental setup, we build upon the open-source FLARE~\cite{kim2025multimodal} framework and replace its Multi-Modal Planner with our model.
We further constrain the planner to output primitive actions only, prohibiting compound actions such as clean, heat, and cool.
For SEEA-R1, it adopts the ALFWorld ‘oracle’ controller. To ensure a fair comparison, we provide FLARE with real-time ground-truth depth and instance segmentation maps from the simulator for low-level execution. 
SEEA-R1 reported a success rate of 46.27\% across 134 tasks in the test set, while our model achieved 60.45\%.


\subsection{Evaluation in Real-Robot Scenarios}

We evaluate our method on real-world robotic scenarios along two dimensions: task planning and task monitoring. First, we construct 50 tabletop scenes with the Agibot A2, each containing up to 9 distinct objects spanning four categories: fruits, beverages, food, and toys. The model is required to plan appropriate pick-and-place actions based on instructions. Second, we construct A2-Failure, a real-world benchmark comprising 160 task-execution trajectories collected on the AgiBot A2. It requires the model to act as an offline failure monitor that determines from video whether a task execution succeeds or fails. 
We compare our model against Qwen3-VL-8B and RoboBrain2.5 on both A2-Failure and tabletop pick-and-place tasks. As shown in Tab.~\ref{tab:real_world}, our model outperforms other baselines on both benchmarks, 
demonstrating the real-world effectiveness of our model. More visual demos are available in the accompanying video.

\section{Conclusion}
\label{sec:conclusion}

In this work, we present \textbf{EmbodiedMind}, trained under an effective paradigm that addresses three key challenges in embodied learning: sample inefficiency, capability imbalance, and long-horizon task planning. Our approach integrates task-aware data curation with difficulty stratification, and introduces Trie-GRPO for step-level advantage estimation via action prefix trees, enabling the policy to distinguish correct intermediate decisions from downstream failures rather than applying uniform trajectory-level penalties. Extensive experiments show that EmbodiedMind achieves outstanding performance on 18 open-source benchmarks. Moreover, we validate its physical grounding in both the simulated environment and real-world robotic deployments, 
demonstrating the effectiveness of our training method.
\begin{table}[htbp]
  \centering
  \vspace{-5pt}
  \caption{Comparison in real-world scenario}
  \label{tab:real_world}
  \begin{tabular}{lccc}
    \toprule
    \textbf{Benchmark} & \textbf{Qwen3-VL-8B} & \textbf{RoboBrain2.5-8B} & \textbf{Ours} \\
    \midrule
    \textbf{A2 Failure}  & 46.88 & 43.75 & 55.31 \\
    \textbf{Pick\&Place} & 40.00  & 42.00  & 62.00  \\
    \bottomrule
  \end{tabular}
  \vspace{-10pt}
\end{table}
\section{Limitation}
\label{sec:Limitations}
Our method relies on difficulty stratification and task-specific rewards. While effective on structured data, extending it to unstructured domains (e.g., general multimodal tasks) is challenging, as categorizing such data and defining robust rewards without fixed formats typically requires large foundation models with prohibitive cost. As shown in Tab.~\ref{tab:time_analy}, rejection sampling on unstructured data incurs substantial time overhead, accounting for 85\% of the RSFT stage. Consequently, we adopted random sampling for these data types, which explains the lack of substantial breakthroughs over baselines on general multimodal benchmarks. Nevertheless, we conducted preliminary experiments to verify the effectiveness of RSFT on unstructured data. Training on 12k Honey-1M samples, RSFT yields no capability degradation compared to the Qwen3-VL-8B across four general multimodal benchmarks, while achieving an average improvement of 0.67\%. Future work will explore more scalable, automated stratification and reward estimation for broader modalities.
\bibliographystyle{IEEEtranBST/IEEEtran}
\bibliography{example}  

\clearpage
\appendices
\section{Datasets}
\label{app:datasets}

We present a comprehensive multimodal training corpus spanning five key categories: general multimodal understanding, 2D/3D spatial perception, video understanding, and task planning. This corpus equips the embodied foundation model with essential capabilities, including world knowledge, commonsense reasoning, and detailed scene understanding. Furthermore, the model's spatiotemporal reasoning—particularly its capacity to interpret fine-grained action semantics—is essential. Building upon this foundation, the model can effectively perform task planning and decomposition in response to complex user instructions.

\subsection{General Multimodal Data}
\label{subsec:general_multimodal}

\textbf{Honey-Data-1M}~\cite{zhang2025bee} is a large-scale multimodal instruction tuning dataset for enhancing logical reasoning and instruction following. We carefully filter out code-mixed data and redundant chain-of-thought samples, retaining only high-quality samples with concise reasoning. Leveraging  Qwen3.5-397B-A17B~\cite{teamqwen3} as an auxiliary classifier, we perform fine-grained categorization into eight distinct classes: basic visual question answering, chart understanding, visual captioning, STEM reasoning, document understanding, visual grounding, object counting, and OCR. This taxonomy spans the perceptual and reasoning capabilities required for embodied intelligence.

\textbf{MM-IFInstruct}~\cite{ding2025mm} aligns textual instructions with visual inputs in a conversational format, spanning image-based dialogue and multimodal perception and generation tasks. Each sample incorporates two types of constraints: \textit{output constraints} (e.g., format requirements, mandatory keywords) and \textit{perception constraints} (e.g., identification, description, or localization of visual elements). This structure requires models to interpret and execute instructions grounded on visual contexts, applicable to embodied scenarios involving scene understanding, spatial reasoning, and object localization.

\subsection{2D Spatial Reasoning}
\label{subsec:spatial_reasoning}

\textbf{EmbSpatial}~\cite{du2024embspatial} is an instruction-tuning dataset designed to enhance the embodied spatial understanding of Visual Language Models (VLMs). We divide its egocentric samples into two tasks: (1) distance estimation — selecting the closest object from a given list; and (2) spatial relation recognition — identifying relative positions such as above, below, left, right, near, and far. The data is then balanced across task types and spatial relations to facilitate robust and unbiased spatial perception.

\textbf{CrossPoint}~\cite{wang2025towards} addresses cross-view spatial consistency (locating corresponding points across different viewpoints). We enforce JSON output format. Furthermore, we retain the multi-dimensional task settings from the original dataset, covering both single-view and cross-view tasks. For single-view tasks, we focus on fine-grained target localization and the understanding of interactable spatial points within indoor scenes. For cross-view tasks, we address core model deficiencies, specifically: (1) cross-view target visibility judgment, which determines whether a target visible in one view remains visible in another; and (2) cross-view point-to-point correspondence, which accurately establishes the mapping of the same physical spatial point across different views.

\textbf{RefSpatial}~\cite{zhou2026roborefer} supports referring expression comprehension and multi-step spatial reasoning in embodied environments. Following the task formulation of RoboBrain-2.5, we consider two core referring tasks: object-level referring and location-level referring. Using heuristic and neural filtering on the original multi-turn QA data, we classify the samples into three categories: direct object referring (e.g., \textit{the white shirt}), spatial localization referring (e.g., \textit{the unoccupied region on the counter in the bottom right}), and relational referring (e.g., \textit{the third object from the left}).

\textbf{CA-VQA-2D}~\cite{daxberger2025mm} is a spatial perception dataset derived from CA-1M~\cite{lazarow2025cubify}, spanning six task categories: existence verification, counting, multiple-choice, distance measurement, point grounding, and 2D/3D visual grounding. Regarding 2D tasks, we subsample judgment, counting, and multiple-choice instances, optimizing prompts to enable multi-frame temporal reasoning. Output formats are standardized for rejection sampling and reinforcement learning. Preprocessing retains multiple-choice samples either without coordinate annotations or with point or bounding box representations.

\textbf{COCO}~\cite{lin2014microsoft} is a large-scale visual grounding dataset. We utilize its object detection annotations to construct a visual grounding QA dataset for enhancing object localization accuracy. To diversify referring expressions and improve grounding generalization, we manually designed 10 prompt templates (e.g., "List the locations of all {labels} you can find in the image using detection boxes") and randomly sampled from them during data generation. For data filtering, we retain categories relevant to indoor scenes, encompassing three types: food and tableware, furniture and indoor items, and portable daily objects. To simplify training, we restrict each sample to at most three object categories and exclude samples containing more than ten bounding boxes.

\subsection{3D Spatial Reasoning}
\label{subsec:3d_spatial}

\textbf{MSMU}~\cite{chen2026sd} provides spatial annotations sourced from 3D scenes, with each image associated with multiple question-answer pairs (without explicit class labels). During preprocessing, we decompose multi-turn dialogues into independent single-turn exchanges to reduce context fragmentation. Using Qwen3.5-397B-A17B, we categorize the data into 2D spatial reasoning and 3D geometric estimation, encompassing eight subcategories: object presence verification, counting, relative positioning, coordinate-based localization, object scale estimation, size comparison, referential size estimation, and absolute distance measurement.

\textbf{CA-VQA-3D} restructures the 3D bounding box and distance data from CA-VQA~\cite{daxberger2025mm} as follows:
\begin{itemize}   
\item \textbf{3D Localization:} We extract dimension components $[l, w, h]$ from the 7D vectors $[cx, cy, cz, l, w, h, yaw]$ and corresponding object references. Prompts are reconstructed to guide the model in regressing 3D dimensions while disregarding coordinate system-sensitive absolute coordinates and yaw. To standardize output, "width" is defined as the longest horizontal dimension (ensuring $w \ge l$), with strict JSON format constraints: \texttt{\{"width": ..., "length": ..., "height": ...\}}.        
\item \textbf{Distance Measurement:} We design two data categories: (1) direct prediction, and (2) chain-of-thought prediction via "reference object identification." For (2), we establish a unit whitelist (m, cm, feet, inch, mm) and perform multiple sampling of training model to collect \textit{responses containing reasoning processes}. Subsequently, we convert both predicted and ground truth distances into centimeters, \textit{selecting the samples with the minimum prediction error (ranging from 0.5 to 2)} as high-quality reasoning training data. These strategies enhance prediction stability under diverse output constraints.
\end{itemize}

\subsection{Task Planning}
\label{subsec:task_planning}

\textbf{SelfReVision}~\cite{park2025making}, \textbf{AgiBot}~\cite{bu2025agibot_iros}, and \textbf{ShareRobot}~\cite{ji2025robobrain} share a unified processing pipeline that converts raw real-world robotic annotations into a standardized format: a natural language \texttt{<response>}, an augmented \texttt{<plans>} field with action tags (e.g., [Navigate], [Manipulate]), and a structured \texttt{<actions>} field represented as action-object-location tuples. All three datasets leverage VLMs as conversion experts, adhering to strict prompting and filtering protocols to preserve semantic integrity. Specifically, for \textbf{SelfReVision}, we enrich GPT-4o-generated plans with action tags and structured action sequences; for \textbf{AgiBot}, we deduplicate task instances, refine task names, and transform annotated atomic steps into \texttt{<plans>} and \texttt{<actions> triples}; and for \textbf{ShareRobot}, starting from only task names and a single frame, we employ a VLM for scene description followed by an LLM to synthesize comprehensive plans and actions. The final output across all three is a unified conversation entry containing images and structured assistant responses, making them highly suitable for robot task planning learning.

\textbf{ALFRED}~\cite{shridhar2020alfred} is an interactive household task planning dataset built on AI2-THOR, comprising 25,743 instructions paired with PDDL expert trajectories. To construct a \textit{spatially grounded} planning corpus, we parse PDDL files for action sequences while capturing panoramic views and object bounding boxes from the simulation. We dynamically track entities using object IDs—binding visible instances to their bounding boxes and resetting them to [] upon pickup or occlusion. Finally, we assign semantic labels (\texttt{[Navigate]}, \texttt{[Map]}, \texttt{[Manipulate]}) based on action types and the availability of bounding boxes.

\subsection{Video Understanding}
\label{subsec:video_understanding}

\textbf{EgoPlan-IT}~\cite{chen2026egoplan} is derived from Epic-Kitchens first-person videos, featuring dense action annotations with temporal boundaries. We design three categories of question-answering pairs: retrospective reasoning (What action was just completed?), proactive planning (What should be done next to achieve the goal?), and multiple-choice verification. Using Qwen3.6-27B, we generate and filter for high-quality training samples by retaining only those where the model generates correct answers supported by coherent reasoning chains.

\textbf{RoboFAC}~\cite{ye2025robofac} is a large-scale video QA dataset for embodied task failure analysis and correction. We adopt the training split collected in ManiSkill simulation~\cite{tao2024maniskill3}. As the raw data lacks task-level annotations, we follow the RoboFAC protocol and use Qwen3.5-397B-A17B to categorize the samples into three types and eight subcategories: (1) \textbf{Task Understanding}: task identification and sub-step planning; (2) \textbf{Failure Analysis}: failure detection, localization, recognition, and explanation; and (3) \textbf{Failure Correction}: high-level replanning and low-level action refinement. For the detection, localization, and recognition subcategories, the original data contains only final outcomes, insufficient for step-wise reasoning supervision. To mitigate this limitation, we use Qwen3.5-397B-A17B to generate step-by-step reasoning chains and filter samples based on reasoning correctness and step completeness, yielding high-quality training data.

\section{Task-specific matching algorithms for Reject sampling}
\label{app:matching-details}
To accommodate diverse output formats across tasks, we introduce a hierarchical matching framework comprising five complementary strategies:

\paragraph{Exact Matching.} 
For multiple-choice questions, true/false judgments, and structured output tasks, we apply exact matching rules. Answers are extracted via regular expressions from \texttt{<answer>} tags and compared against ground truth through strict string equality.

\paragraph{Point Localization Matching.} 
For point localization tasks, coordinates are first normalized to the $[0, 1000]$ range, and errors are measured using L1 distance. Given a predicted point $\mathbf{p}_p = (x_p, y_p)$ and ground truth point $\mathbf{p}_g = (x_g, y_g)$, the matching function is defined as:
\begin{equation}
    \mathcal{M}(\mathbf{p}_p, \mathbf{p}_g) = \mathbb{I}\big( |x_p - x_g| + |y_p - y_g| \leq \tau_d \big),
\end{equation}
where $\tau_d$ denotes the distance threshold.

\paragraph{Spatial Distance Estimation Matching.} 
For spatial distance estimation tasks, all predicted distances are first normalized to centimeters. Since distance regression errors are inherently bidirectional, both underestimation and overestimation must be penalized equally. To address this, we introduce a ratio consistency criterion. Given a predicted distance $d_p$ and ground truth distance $d_g$, the ratio $R$ is computed as:
\begin{equation}
    R = \min\left(\frac{d_p}{d_g}, \frac{d_g}{d_p}\right),
    \label{eq:distance_ratio}
\end{equation}
where $R \leq 1$ by construction. A prediction is considered correct when $R \geq \tau_r$, with $\tau_r$ denoting the predefined threshold.

\paragraph{3D Object Size Estimation.} 
For 3D object size regression tasks, we compute the 3D Intersection-over-Union (IoU). Denoting the predicted box volume as $V_p$, ground truth volume as $V_g$, and intersection volume as $V_{\text{inter}}$, the IoU is defined as:
\begin{equation}
    \text{IoU}_{\text{3D}} = \frac{V_{\text{inter}}}{V_p + V_g - V_{\text{inter}}},
    \label{eq:3d_iou}
\end{equation}
Predictions satisfying $\text{IoU}_{\text{3D}} \geq \tau_{\text{iou}}$ are deemed correct, where $\tau_{\text{iou}}$ is the IoU threshold.
    
\paragraph{LLM-Based Semantic Verification.} 
While rule-based methods offer high efficiency and determinism, they may fail on complex, unstructured natural language outputs where semantic correctness diverges from rigid criteria. To address this, we introduce an LLM-based judge as a complementary verification layer. Both the model-generated answer and ground truth are fed into Qwen3.5-397B-A17B, which performs semantic alignment analysis via a carefully designed Chain-of-Thought prompt. This approach tolerates variations in phrasing, additional contextual details, or alternative sentence structures, provided that core factual content remains preserved.

\section{Multi-Task Hybrid Reward Mechanisms}

\label{app:reward-design}
For closed-form tasks such as multiple-choice questions, binary judgments, and spatial point localization, we adopt the matching functions defined in Section~\ref{app:matching-details} to obtain binary rewards. For 3D distance regression tasks, we use the distance ratio from Eq.~\ref{eq:distance_ratio} as the reward signal. For other tasks, we devise specialized reward functions as follows:

\paragraph{Multi-Object Hungarian Matching Reward.}
In multi-object detection, predicted bounding boxes $\mathcal{P} = \{\mathbf{b}_1^p, \dots, \mathbf{b}_M^p\}$ lack a predefined correspondence with ground-truth boxes $\mathcal{G} = \{\mathbf{b}_1^g, \dots, \mathbf{b}_N^g\}$, and the cardinalities $M$ and $N$ frequently differ. To evaluate predictions under this ambiguity, we formulate a reward using the Hungarian matching algorithm.

We first compute pairwise IoU between $\mathcal{P}$ and $\mathcal{G}$, constructing an $N \times M$ similarity matrix $\mathbf{A}$. Each entry $A_{i,j}$ quantifies the geometric overlap between ground-truth box $\mathbf{b}_i^g$ and predicted box $\mathbf{b}_j^p$:
\begin{equation}
    A_{i,j} = \operatorname{IoU}(\mathbf{b}_i^g, \mathbf{b}_j^p) = \frac{\operatorname{Area}(\mathbf{b}_i^g \cap \mathbf{b}_j^p)}{\operatorname{Area}(\mathbf{b}_i^g \cup \mathbf{b}_j^p)},
\end{equation}

To establish optimal correspondence between predictions and ground truth, we apply the Hungarian algorithm to obtain the matching index set $\mathcal{K}$ by minimizing the sum of negative IoUs. After matching, we retain only pairs satisfying $A_{i,j} \geq \tau$ as valid matches, denoting the count of valid matches as $N_{\text{valid}}$. The detection reward combines the $F_1$ score with matching quality to penalize false negatives and false positives while capturing localization accuracy:

\begin{equation}
    R_{\text{det}} = \underbrace{\frac{2 \cdot P \cdot R}{P + R}}_{F_1 \text{ score}} \times \underbrace{\left( \frac{1}{N_{\text{valid}}} \sum_{(i,j) \in \mathcal{K}, A_{i,j} \geq \tau} A_{i,j} \right)}_{\text{mean IoU of valid matches}}
\end{equation}

\paragraph{3D Object Size Regression Reward.}
For 3D object size estimation, we design a center-aligned 3D Volume IoU reward to measure deviations between predicted and ground-truth dimensions. Under the assumption that predicted and ground-truth boxes share aligned geometric centers and parallel axes in 3D space, the reward equals the ratio of intersection volume to union volume. Denoting ground-truth dimensions as $(w_g, l_g, h_g)$ and predicted dimensions as $(w_p, l_p, h_p)$, we define:
\begin{equation}
    R_{\text{size}} = \frac{V_{\text{inter}}}{\max(V_{\text{union}}, \epsilon)},
\end{equation}
where $\epsilon$ prevents numerical instability. The intersection and union volumes are computed as:
\begin{align}
    V_{\text{inter}} &= \min(w_p, w_g) \cdot \min(l_p, l_g) \cdot \min(h_p, h_g), \\
    V_{\text{union}} &= (w_p \cdot l_p \cdot h_p) + (w_g \cdot l_g \cdot h_g) - V_{\text{inter}}.
\end{align}

\section{Trie-based Group Relative Policy Optimization}
\subsection{Output Format Specifications for Policy and Reward Models}
\label{app:Trie-GRPO:plan-format}

To facilitate automated parsing and reliable credit assignment, we enforce strict XML-structured output protocols for both the policy and reward models. The specifications are detailed below.

\paragraph{Policy Model Output.}
The policy model $\pi_\theta$ generates a structured plan consisting of three mandatory fields. An illustrative example is provided below:


\begin{verbatim}
<response>Okay, I will put the pan with 
the spatula in the sink.</response>
<plans>
1.[Navigate] Navigate to the Spatula.
2.[Manipulate] Pick up the Spatula.
...
</plans>
<actions>
[['Navigate', 'Spatula'], 
['Pick', 'Spatula'], ...]
</actions>
\end{verbatim}

\paragraph{Reward Model Output.}
The reward model $R_\phi$ evaluates the generated plan and outputs a binary judgment accompanied by a structured rationale. An example is shown below:


\begin{verbatim}
<Reason>The task requires placing a heated 
piece of potato into the fridge. The plan 
includes slicing, a complete microwave 
heating process, and successfully placing 
the heated potato slice into the fridge. 
The manipulated objects (Potato, 
ButterKnife, Fridge, Microwave) are
visible in the field of view. The steps 
are complete, and each planning step is 
logically correct.</Reason>
<Results>Success</Results>
\end{verbatim}


\subsection{Tree-Guided Training Signal Generation}
\label{app:Trie-GRPO:training-signal}

\begin{algorithm}
\caption{Trie-GRPO Training Signal Generation}
\label{alg:trie-grpo}
\SetAlgoLined
\SetAlgoSkip{smallskip}
\KwIn{Query $q$, policy $\pi_\theta$, reward model $R$, discount $\gamma$, samples $K$, votes $M$}
\KwOut{Advantage-annotated dataset $\mathcal{D}$}
\BlankLine
\BlankLine
\noindent\textbf{Stage 1--2: Trajectory Sampling and Reward Labeling}

Sample $K$ trajectories from Query q, label rewards via reward model's $M$-vote\;
$\{\tau_i\}_{i=1}^K \sim \pi_\theta(q)$\;
$R_i \leftarrow \text{sign} ( \sum_{m=1}^M R^{(m)}(\tau_i) )$\;
\BlankLine
\BlankLine
\noindent\textbf{Stage 3: Action Parsing and Trie Construction}

Parse and build action prefix tree\;
$\mathcal{T} \leftarrow \text{BuildTrie}\big(\{\text{parse}(\tau_i)\}_{i=1}^K\big)$\;
\BlankLine
\BlankLine
\noindent\textbf{Stage 4: Discounted Reward Backpropagation and Q-Value Estimation}

Backpropagate discounted rewards along paths\;
\ForEach{$\tau_i \in \mathcal{T}$}{
    \ForEach{$v_t \in \text{path}(\tau_i)$}{
        $v_t.\text{rewards} \mathrel{+}= \gamma^{T_i-t} \cdot R_i$\;
    }
}
\BlankLine
Compute Q-values\;
$Q(v) \leftarrow \frac{1}{|v.\text{rewards}|}\sum_{r \in v.\text{rewards}} r$, ~~ $\forall v \in \mathcal{T}$\;
\BlankLine
\BlankLine
\noindent\textbf{Stage 5-6: Sibling-Normalized Advantage Computation and Low-Variance Sample Filtering}

Compute advantages within sibling groups\;
$\mathcal{D} \leftarrow \emptyset$\;
\ForEach{parent $u \in \mathcal{T}$}{
    $\mu_u \leftarrow \frac{1}{k}\sum_{j=1}^{k} Q(v_j)$\;
    $\sigma_u \leftarrow \sqrt{\frac{1}{k}\sum_{j=1}^{k}(Q(v_j)-\mu_u)^2} + \epsilon$\;
    \ForEach{$v_j \in \text{children}(u)$}{
        $A_j \leftarrow (Q(v_j) - \mu_u) / \sigma_u$\;
        \If{$|A_j| > 0.001$}{
            $\mathcal{D} \mathrel{+}= \{(\text{context}(v_j), \text{action}(v_j), A_j)\}$\;
        }
    }
}
\BlankLine
\BlankLine

\Return $\mathcal{D}$\;
\end{algorithm}

\begin{algorithm}
\caption{Trie-GRPO Automated Iterative Training Framework}
\label{alg:framework}
\SetAlgoLined
\SetAlgoSkip{smallskip}
\KwIn{Policy $\pi_{\theta_0}$, reward model $R_{\phi_0}$, reference $\pi_{\text{ref}}$, max rounds $R_{\max}$, tasks $\mathcal{Q}$, batches $\{\text{split}_b\}$, discount $\gamma$, reward model SFT dataset $D_{pre}$}
\KwOut{Final policy $\pi_{\theta_R}$}

\BlankLine
\BlankLine

\noindent\textbf{Round 1 to $R_{\max}$: Iterative Training}

\For{$r \leftarrow 1$ \KwTo $R_{\max}$}{
    \noindent\textbf{Stage 1: Data Generation}
    
    $\mathcal{Q}_r \leftarrow \text{split}_{r \bmod B}$\;
    $\mathcal{D}_r \leftarrow \emptyset$\;
    \ForEach{task $q \in \mathcal{Q}_r$}{
        $\{\tau_k\}_{k=1}^K \leftarrow \pi_{\theta_{r-1}}(q)$\;
        \ForEach{$\tau_i \in \{\tau_k\}$}{
            $R_i \leftarrow \text{majority\_vote}\big(\{R_{\phi_{r-1}}(q, \tau_i)^{(m)}\}_{m=1}^M\big)$\;
        }
        $\mathcal{D}_r^{(q)} \leftarrow \text{Algorithm~\ref{alg:trie-grpo}}(\{(\tau_i, R_i)\}, \gamma)$\;
        $\mathcal{D}_r \leftarrow \mathcal{D}_r \cup \mathcal{D}_r^{(q)}$\;
    }
    \BlankLine
    \BlankLine
    
    \noindent\textbf{Stage 2: Logps Preprocessing}
    
    ref\_logps $\leftarrow \pi_{\text{ref}}$.logprobs($\mathcal{D}_r$)\;
    old\_logps $\leftarrow \pi_{\theta_{r-1}}$.logprobs($\mathcal{D}_r$)\;
    \BlankLine
    \BlankLine
    
    \noindent\textbf{Stage 3: Policy Model Update}
    
    $\pi_{\theta_r} \leftarrow \text{GRPO\_Train}(\pi_{\theta_{r-1}}, \mathcal{D}_r, \text{ref\_logps}, \text{old\_logps})$\;
    \BlankLine
    \BlankLine
    
    \noindent\textbf{Stage 4: Reward Model DAPO Training}
    
    $\mathcal{D}_r^{\text{bal}} \leftarrow \text{resample}(\mathcal{D}_r, \text{Success/Failure ratio}=1:1)$\;
    $R_{\phi_r'} \leftarrow \text{DAPO\_Train}(R_{\phi_{r-1}}, \mathcal{D}_r^{\text{bal}})$\;
    \BlankLine
    \BlankLine
    
    \noindent\textbf{Stage 5: Reward Model SFT (Every 2 Rounds)}
    
    \eIf{$r \bmod 2 = 0$}{
        $R_{\phi_r} \leftarrow \text{SFT\_Train}(R_{\phi_{r-1}}, \mathcal{D}_{\text{pre}})$\;
    }{
        $R_{\phi_r} \leftarrow R_{\phi_r'}$\;
    }
    \BlankLine
    \BlankLine
    \noindent\textbf{Stage 6: Early Stopping (Optional)}
    
    \If{val\_perf unchanged for $k$ rounds}{
        Break;
    }
    \BlankLine
    \BlankLine
}
\KwRet{$\pi_{\theta_r}$}\;
\end{algorithm}

This section presents the complete algorithmic pipeline for Trie-GRPO training signal generation. Given an input query $q$, the algorithm transforms raw rollout trajectories into advantage-annotated training data through six stages: (1) multi-candidate trajectory sampling, (2) voting-based reward labeling, (3) action parsing and prefix tree construction, (4) temporal-discounted reward propagation and Q-value estimation, (5) sibling-normalized advantage computation, and (6) low-variance sample filtering and data formatting. The complete Trie-GRPO training signal generation procedure is summarized in Algorithm~\ref{alg:trie-grpo}.

\begin{table*}[h]
\caption{Evaluation results of the EmbodiedMind-27B model on different benchmarks. (Abbreviations — Rynn-30B: RynnBrain-30B-A3B~\cite{dang2026rynnbrain}, Pelican-72B: Pelican1.0-VL-72B~\cite{zhang2025pelican}, Qwen3.5-397B: Qwen3.5-397B-A17B-FP8)}
\label{tab:embodiedmind27b_results}
\centering
\begin{tabular}{cccccc}
\hline
\textbf{Benchmark}                       & \textbf{Qwen3.6-27B} & \textbf{Rynn-30B} & \textbf{Pelican-72B} & \textbf{Qwen-397B} & \textbf{Ours--27B} \\ \hline
\multicolumn{6}{c}{\textbf{General Benchmarks}}                                                                                                      \\ \hline
AI2D~\cite{kembhavi2016diagram}          & 88.92                & 85.78             & 86.43                & 90.77              & 85.40              \\
MMSTAR~\cite{chen2024we}                 & 71.67                & 68.53             & 69.07                & 79.27              & 70.27             \\
MMMU~\cite{yue2024mmmu}                  & 78.29                & 58.75             & 61.47                & 80.48              & 70.67            \\
OCRBench~\cite{liu2024ocrbench}          & 76.80                & 72.90             & 78.50                & 82.70              & 72.10              \\
\textbf{Average}                         & \textbf{78.92}       & \textbf{71.49}    & \textbf{73.87}       & \textbf{83.31}     & \textbf{74.61}     \\ \hline
\multicolumn{6}{c}{\textbf{2D Reasoning}}                                                                                                            \\ \hline
CV-Bench~\cite{tong2024cambrian}         & 86.80                & 89.49             & 83.50                & 87.98              & 88.22              \\
CrossPoint~\cite{wang2025towards}        & 40.00                & 38.50             & 36.60                & 53.40              & 78.20              \\
RoboSpatial~\cite{song2025robospatial}   & 64.86                & 74.00             & 58.57                & 74.86              & 69.43              \\
RefSpatial~\cite{zhou2026roborefer}      & 25.00                & 57.50             & 48.50                & 62.50              & 67.00              \\
EMbSpatial~\cite{du2024embspatial}       & 82.64                & 81.76             & 76.95                & 83.30              & 82.77              \\
ERQA~\cite{team2025gemini}               & 49.00                & 44.25             & 43.25                & 50.50              & 51.50              \\
\textbf{Average}                         & \textbf{58.05}       & \textbf{64.25}    & \textbf{57.90}       & \textbf{68.76}     & \textbf{72.85}     \\ \hline
\multicolumn{6}{c}{\textbf{3D Measurement}}                                                                                                          \\ \hline
MSMU~\cite{chen2026sd}                   & 32.93                & 51.69             & 54.58                & 31.53              & 84.45              \\
BLINK~\cite{fu2024blink}                 & 85.31                & 85.31             & 80.42                & 86.01              & 83.92              \\
Q-Spatial~\cite{liao2024reasoning}       & 62.36                & 46.86             & 28.78                & 71.22              & 63.84              \\
\textbf{Average}                         & \textbf{60.20}       & \textbf{61.29}    & \textbf{54.59}       & \textbf{62.92}     & \textbf{77.40}     \\ \hline
\multicolumn{6}{c}{\textbf{Planning}}                                                                                                                \\ \hline
ShareRobot-Judge~\cite{ji2025robobrain}  & 96.05                & 70.70             & 97.75                & 97.75              & 95.45              \\
ShareRobot-Choice~\cite{ji2025robobrain} & 90.16                & 84.52             & 84.56                & 89.73              & 88.15              \\
EgoPlan-1~\cite{chen2026egoplan}         & 59.35                & 53.89             & 56.10                & 58.97              & 65.48              \\
EgoPlan-2\cite{qiu2026egoplan}           & 53.84                & 53.38             & 54.37                & 51.94              & 58.56              \\
RoboBench~\cite{luo2025robobench}        & 57.86                & 35.91             & 37.07                & 59.41              & 53.57              \\
\textbf{Average}                         & \textbf{71.45}       & \textbf{59.68}    & \textbf{65.97}       & \textbf{71.56}     & \textbf{72.24}     \\ \hline
\end{tabular}%
\end{table*}

\begin{figure}[t]
    \centering
    \includegraphics[width=\linewidth]{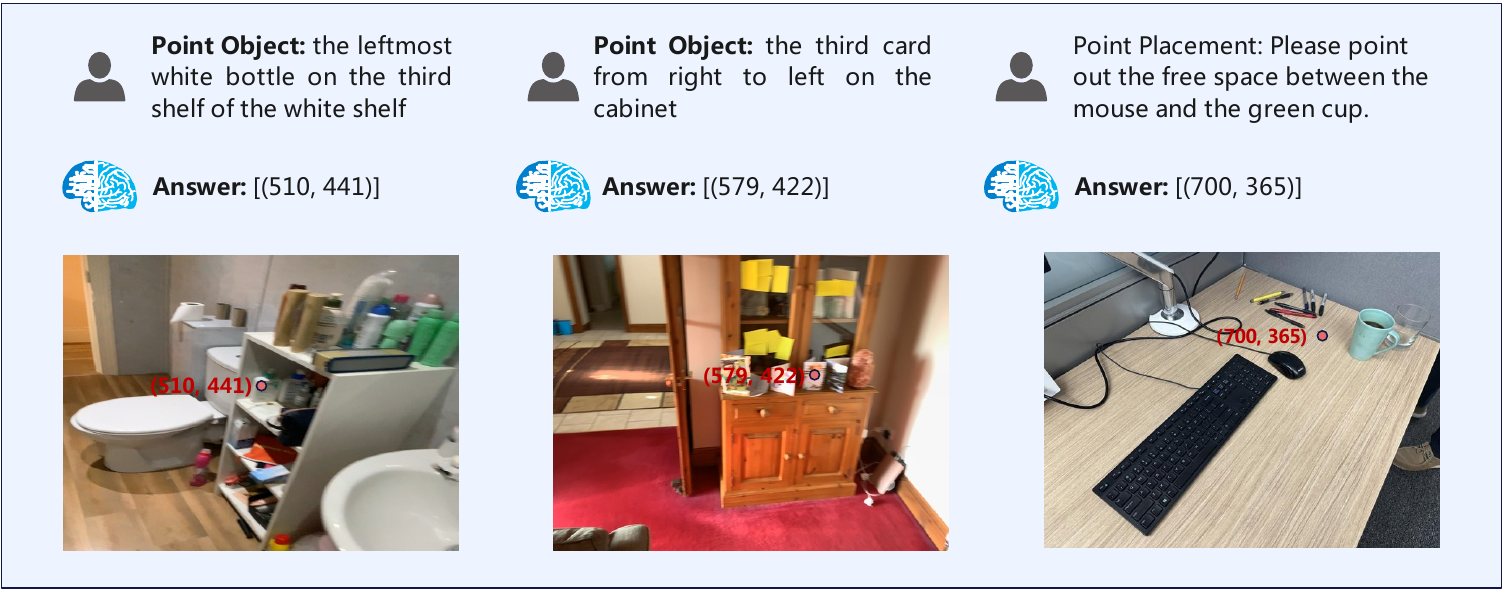}
    \vspace{-15pt}
    \caption{Visualization of Point Localization tasks in the Refspatial benchmark.}
    \label{fig:spatial_reasoning_vis_Ref}
\end{figure}

\begin{figure}[t]
    \centering
    \includegraphics[width=\linewidth]{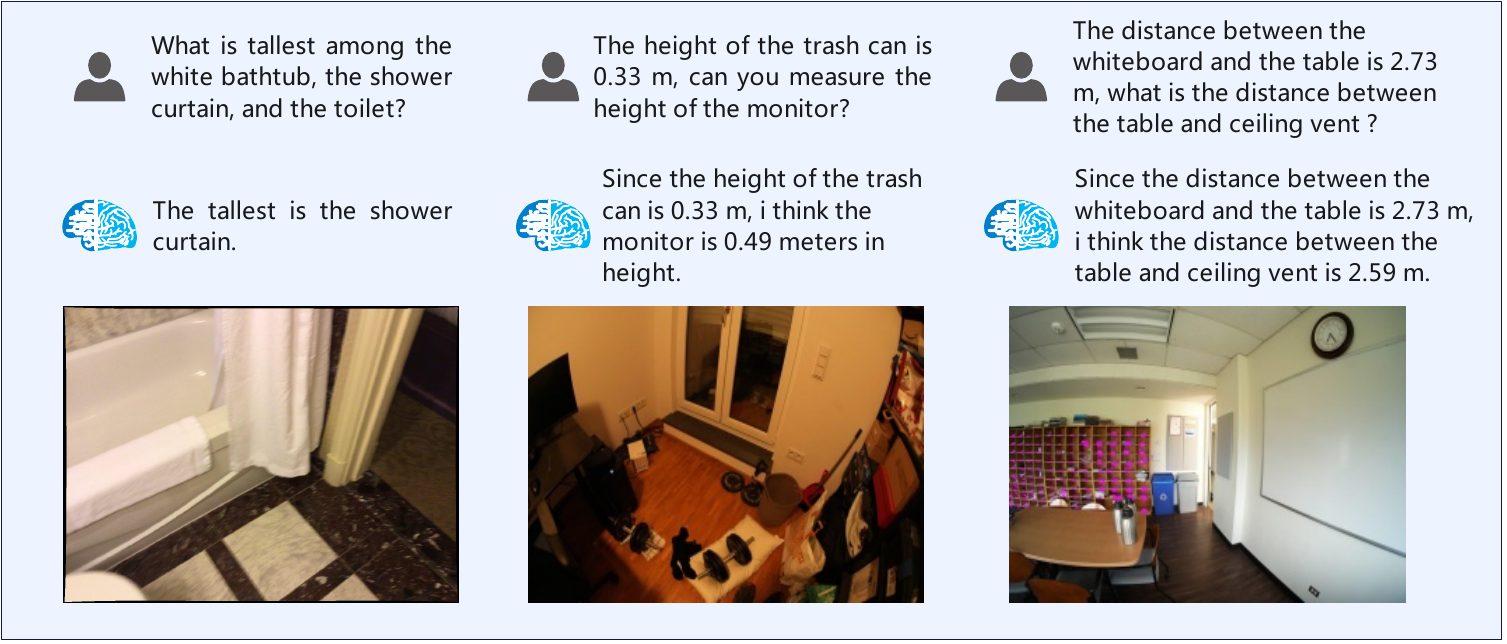}
    \vspace{-15pt}
    \caption{Visualization of 3D  Measurement tasks in the MSMU benchmark.}
    \label{fig:spatial_reasoning_vis_MSMU}
\end{figure}

\begin{figure*}[t]
    \centering
    \includegraphics[width=\linewidth]{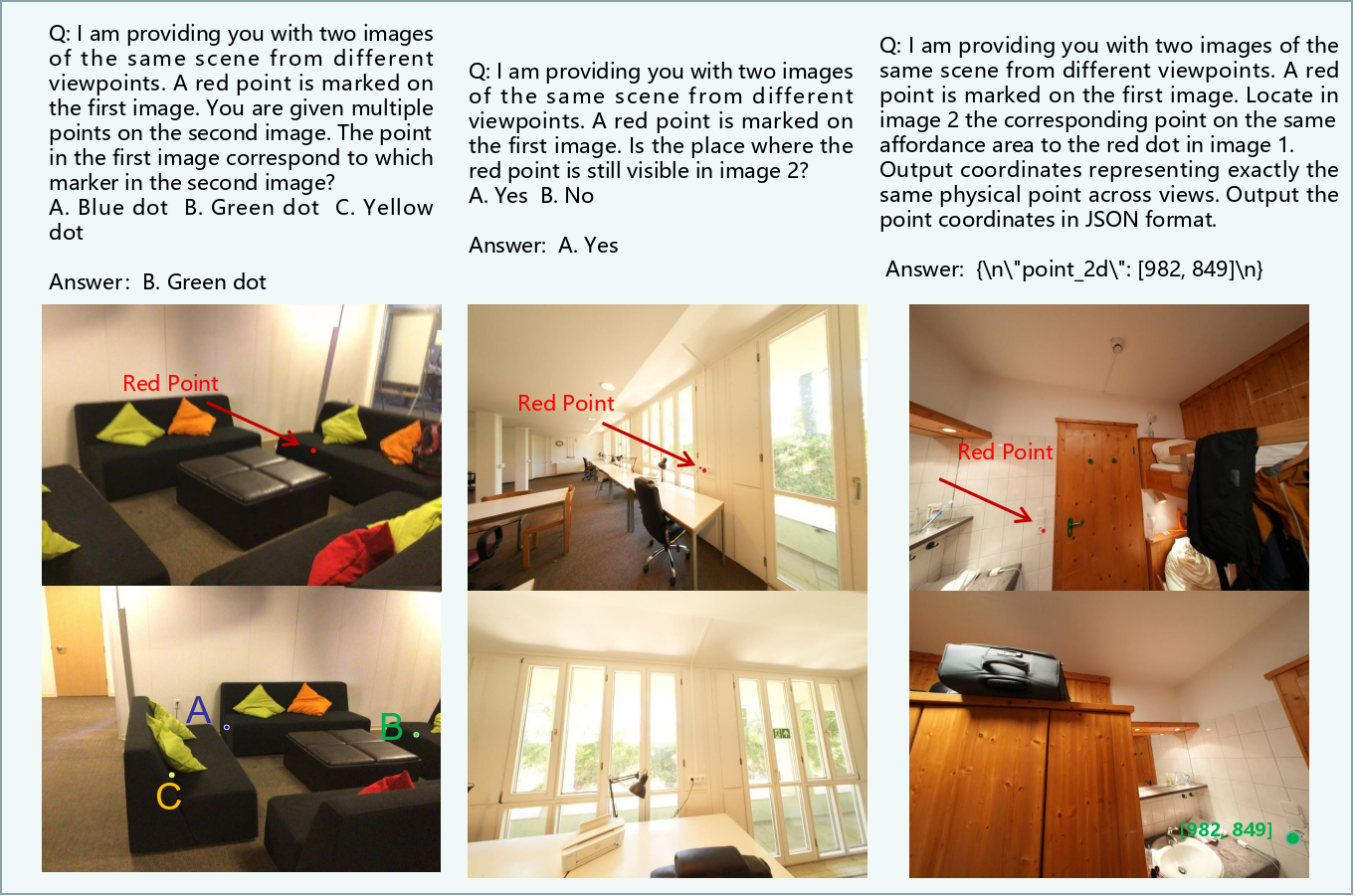}
    \vspace{-15pt}
    \caption{Visualization of Cross-view point localization tasks in the CrossPoint benchmark.}
    \label{fig:spatial_reasoning_vis_Cross}
\end{figure*}

\subsection{Automated Iterative Training Algorithm for Trie-GRPO}
\label{app_alg:trie-grpo-steps}

The entire training process comprises $R$ iterations. \textbf{Round 1 (Cold Start)}: If the user provides preset high-quality demonstration data, skip the sampling phase and use it directly for training; otherwise, perform sampling using the initial policy model to generate training data. \textbf{Round 2 and Beyond}: Starting from Round 2, the framework enters the self-iteration phase. To control computational costs per round, we partition the task set $\mathcal{Q}$ into $B$ batches, executing the complete data generation pipeline on only one batch per round. Specifically, the task subset processed in round $r$ is $\mathcal{Q}_r = \text{split}_{r \bmod B}$. This design resembles curriculum learning, enabling the model to focus on different task subsets across rounds while mitigating catastrophic forgetting.

Each round $r$ (for $r \geq 2$) executes the following six stages, as summarized in Algorithm~\ref{alg:framework}:

(1) \textbf{Batch Rotation and Data Generation}: To control computational costs, we partition the task set $\mathcal{Q}$ into $B$ batches and process one batch per round via $\mathcal{Q}_r = \text{split}_{r \bmod B}$. For each task $q \in \mathcal{Q}_r$, we perform $K$ rollouts using $\pi_{\theta_{r-1}}$ and $M$-vote reward labeling, then invoke Algorithm~\ref{alg:trie-grpo} to generate advantage-annotated dataset $\mathcal{D}_r$.

(2) \textbf{Logps Preprocessing}: We compute per-token log probabilities from both the reference model $\pi_{\text{ref}}$ and the previous policy $\pi_{\theta_{r-1}}$, which are used for KL-divergence constraint and importance sampling, respectively.

(3) \textbf{Policy Model Update}: The policy model is optimized via GRPO using $\mathcal{D}_r$ along with the precomputed logps, yielding $\pi_{\theta_r}$.

(4) \textbf{Reward Model GRPO Training}: We fine-tune the reward model using pairwise ranking loss on trajectory-reward pairs from $\mathcal{D}_r$, with balanced positive/negative sampling.

(5) \textbf{Reward Model SFT (every 2 rounds)}: On even-numbered rounds ($r = 2, 4, 6, \ldots$), we additionally perform supervised fine-tuning using high-confidence samples (voting consistency $\geq 4/5$) to consolidate the reward model's judgment capability.

(6) \textbf{Early Stopping Check (optional)}: Training terminates if validation performance shows no significant improvement for $k$ consecutive rounds or reaches the preset maximum number of rounds.

\section{Evaluation results of the EmbodiedMind-27B}
\label{app:additional_results}
We further scale up our RSFT to EmbodiedMind-27B, initialized from Qwen3.6-27B~\cite{teamqwen3.6}. Results are presented in Table~\ref{tab:embodiedmind27b_results}, with comprehensive comparisons against stronger baselines including Pelican-VL-72B and RynnBrain-30B-A3B.

\section{Visualization}
\subsection{Spatial Reasoning Visualization}
To validate the spatial reasoning capabilities of our model, we present qualitative visualizations of EmbodiedMind-8B across multiple benchmarks, as shown in Figures~\ref{fig:spatial_reasoning_vis_Ref}, \ref{fig:spatial_reasoning_vis_MSMU}, and \ref{fig:spatial_reasoning_vis_Cross}.

\subsection{Long-Horizon Planning Visualization}
\begin{figure*}[t]
    \centering
    \includegraphics[width=\linewidth]{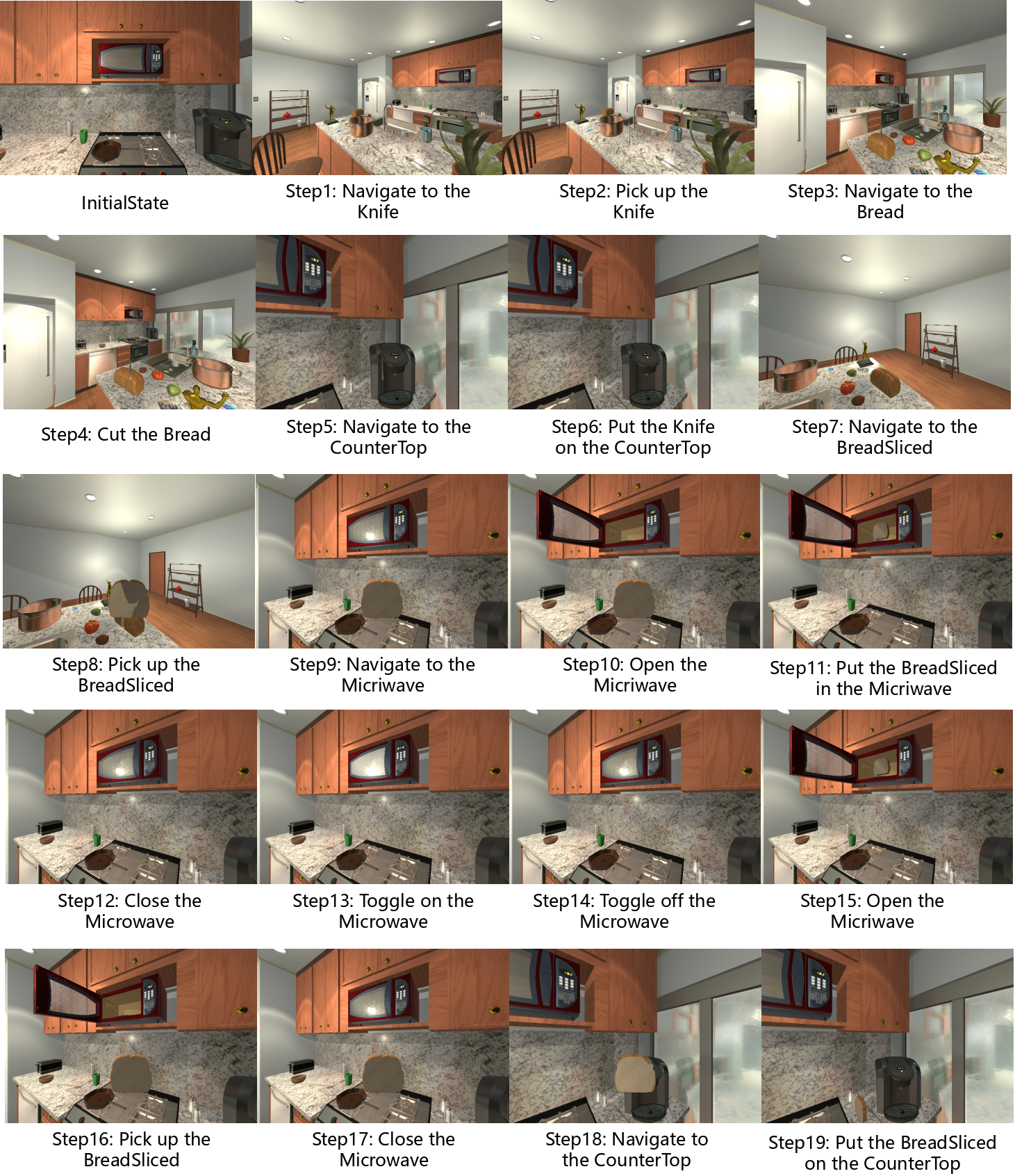}
    \vspace{-5pt}
    \caption{Visualization of long-horizon planning tasks in the AI2-THOR simulated environment; Task: Cut the bread, heat the slice of bread in the microwave, and then place the slice on the counter.}
    \label{fig:long_horizon_vis}
    \vspace{-10pt}
\end{figure*}

To evaluate long-horizon planning, we select three tasks from VLM-PlanSim-99~\cite{zou2025embodiedbrain}, each requiring approximately 20 steps. We visualize one complete execution trace in AI2-THOR, as shown in Figure~\ref{fig:long_horizon_vis}. Additionally, the execution videos for all three tasks are provided in the attached zip file.

\subsection{Real-World Task Planning Visualization}
\begin{figure*}[t]
    \centering
    \includegraphics[width=\linewidth]{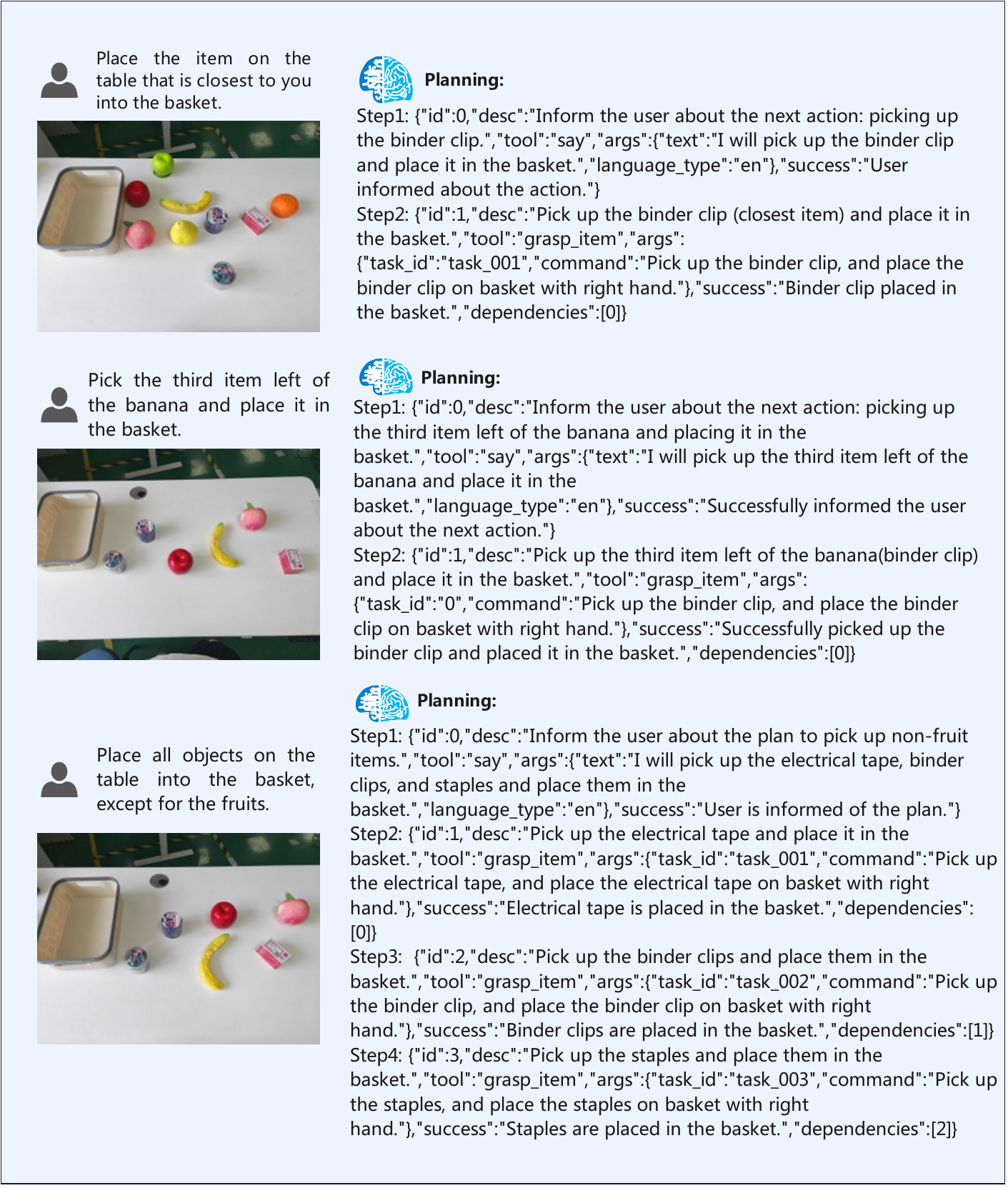}
    \vspace{-5pt}
    \caption{Real-World Task Planning Visualization.}
    \label{fig:real_world_planning}
\end{figure*}

To validate the task planning capabilities of our model beyond simulation, we deploy it on an embodied humanoid robot and evaluate its performance in real-world environments. As illustrated in Figure~\ref{fig:real_world_planning}, the model demonstrates robust spatial reasoning, accurately inferring object relationships and decomposing tasks into executable subgoals with precise tool selection. Supplementary videos of complete task executions are provided in the attached materials.

\end{document}